# Advances in Computer-Aided Diagnosis of Diabetic Retinopathy

Saket S. Chaturvedi*, Kajol Gupta, Vaishali Ninawe, Prakash S. Prasad*
Department of Computer Science & Engineering, Priyadarshini Institute of Engineering &
Technology, Nagpur, India
Email: saketschaturvedi@gmail.com, prakashsprasad@gmail.com

**ABSTRACT**

Diabetic Retinopathy is a critical health problem influences 100 million individuals worldwide, and these figures are expected to rise, particularly in Asia. Diabetic Retinopathy is a chronic eye disease which can lead to irreversible vision loss. Considering the visual complexity of retinal images, the early-stage diagnosis of Diabetic Retinopathy can be challenging for human experts. However, Early detection of Diabetic Retinopathy can significantly help to avoid permanent vision loss. The capability of computer-aided detection systems to accurately and efficiently detect the diabetic retinopathy had popularized them among researchers. In this review paper, the literature search was conducted on PubMed, Google Scholar, IEEE Explorer with a focus on the computer-aided detection of Diabetic Retinopathy using either of Machine Learning or Deep Learning algorithms. Moreover, this study also explores the typical methodology utilized for the computer-aided diagnosis of Diabetic Retinopathy. This review paper is aimed to direct the researchers about the limitations of current methods and identify the specific areas in the field to boost future research.

## 1. INTRODUCTION

Diabetic Retinopathy is a critical health problem, which currently influences 100 million individuals worldwide. The even devastating is that visual impairment and blindness estimates for Diabetic Retinopathy increased by 64% and 27% between 1990 and 2010, respectively [1]. Diabetic retinopathy (DR) is the most common complication of diabetes [2]. Since 1980, the cases of diabetes prevalence have quadrupled, and 422 million adults are affected by either form of diabetes, and this number is expected to rise rapidly, particularly in Asia [3], [4].

Diabetic Retinopathy is a chronic eye disease which can lead to irreversible vision loss [5]. Diabetic Retinopathy can be categorized into four stages: mild, moderate, severe nonproliferative diabetic retinopathy, and advanced proliferative diabetic retinopathy. The proliferative diabetic retinopathy is the most dangerous stage which increases the likelihood of blood leaking, causing permanent vision loss. The current methodology of medical imaging for diabetic retinopathy is invasive and unpleasant. Fundus Photography is the most typical procedure for the detection of eye disorder in a diabetic patient by analyzing three retinopathy symptoms: Exudates, Hemorrhages, and Microaneurysms. Early stages of Diabetic Retinopathy can be challenging to diagnose, considering the minimal visible signs [6].

This review paper is organized as follows. Section 2 covers the literature review. Section 3 discusses the methodology of this study, including the typical methodology for the computer-aided diagnosis of Diabetic Retinopathy. Section 4 discusses the identified areas in this study for future research. Section 5 concludes the paper.

## 2. LITERATURE REVIEW

The capability of computer-aided detection systems to accurately and efficiently detect the diabetic retinopathy had made them popular among researchers. Last ten years have recorded numerous research work focusing on the development of Automated Diabetic Retinopathy detection using traditional machine learning algorithms.

Quellec et al. [7] used a KNN algorithm with optimal filters on two classes to achieve an AUC of 0.927. Also, Sinthanayothin et al. [8] proposed an automated Diabetic Retinopathy detection system using KNN algorithm on morphological features to obtain sensitivity and specificity of 80.21% and 70.66%, respectively. Further, in paper [9], three classes of diabetic retinopathy were classified using NN. They classified mild, moderate, and severe stages with an accuracy of 82.6%, 82.6%, and 88.3% respectively.

Larsen et al. [10] demonstrated an automatic diagnosis of Diabetic Retinopathy in fundus photographs with visibility threshold. Their method reported an accuracy of 90.1% for true cases detection and 81.3% for the detection of the false case. Agurto et al. [11] utilized multi-scale Amplitude Modulation and Frequency Modulation based decomposition to distinguish between Diabetic Retinopathy and normal retina images. In [12], the authors reported an area under ROC of 0.98 for Texture features and accuracy of 99.17% for two-class classification by using Wavelet transform with SVM.

Jelinek et al. [13] proposed an automated Diabetic Retinopathy detection by combining the works of Spencer [14] and Cree [13] system, which achieved a sensitivity of 85% and specificity of 90%. Abràmo et al. [15] developed Eye-Check algorithm for automated Diabetic Retinopathy detection. They detected abnormal lesions with an AUC of 0.839. Dupas et al. [16] developed a Computer-Aided Detection system with k-NN classifier to detect Diabetic Retinopathy with a sensitivity of 83.9% and specificity of 72.7%.

Acharya et al. [17] classified five classes using SVM classifier on the bi-spectral invariant features to achieve sensitivity, specificity, and accuracy of 82%, 86%, and 85.9%, respectively. They also worked utilizing four features and achieved a classification accuracy of 85%, sensitivity of 82%, and specificity of 86%. Roychowdhury et al. [18] proposed a two-step classification approach. In the first step, the false positives were removed. Later, GMM, KNN, and SVM were utilized for the classification task. They reported a sensitivity of 100%, specificity of 53.16%, and AUC of 0.904.

Deep learning algorithms have become popular in the last few years. Kaggle [19] has launched several competitions focusing on automated Diabetic Retinopathy detection. Pratt et al. [20] introduced a CNN based method, which even surpassed human experts in the classification of advanced stage Diabetic Retinopathy. Kori et al. [21] utilized an ensemble of ResNet and Densely connected networks to detect advanced stages of Diabetic Retinopathy and macular enema. Torrey et al. [22] developed more interpretable CNN model to detect lesion in the retinal fundus images. In similar study [23], further advancement in the automated diabetic retinopathy methods was done along with RAM. Yang et al. [24] employed unbalanced weight map methodology to emphasize lesion detection with an AUC of 0.95. In [25], VGG-16 and Inception- 4 networks were utilized for effective Diabetic Retinopathy classification.

## 3. METHODOLOGY

The literature search was conducted on PubMed, Google Scholar, IEEE Explorer with no constraint on language or year of publication. The review focuses on the computer-aided detection of Diabetic Retinopathy using either of Machine Learning or Deep Learning algorithms. Moreover, the possible future research areas in the field are proposed to boost future research for the automated detection of diabetic retinopathy and direct the researchers about the challenges and limitations of current methods.

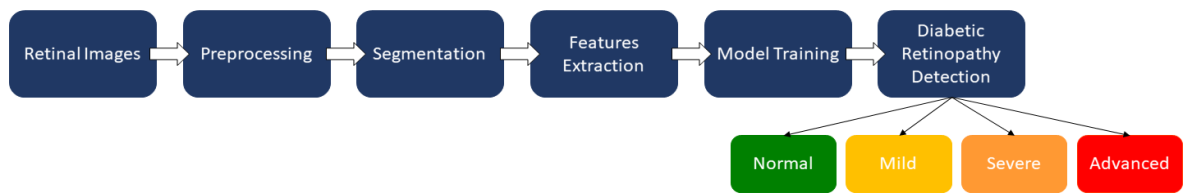

Figure 1. A typical method for automated Diabetic Retinopathy detection

The typical method for automated Diabetic Retinopathy detection covers Preprocessing, Segmentation, Diabetic Retinopathy Features Extraction, Model training, and Diabetic Retinopathy Detection, as illustrated in Figure 1. The starting step is the preprocessing, where the retinal images are pre-processed to enhance and optimize image characteristics. The next step is retinal image segmentation is performed to extract different features from the image for the task. Further, the retinal image dataset is divided into training and test data. The model is trained on the training data to learn the task-specific features. The model training involves tuning of hyperparameters such as batch size, epoch, metrics, etc. Finally, the model performs classification of retinal images, and the performance of model is evaluated over test data employing the metrics like Accuracy, Sensitivity, Specificity, AUC, etc.

## 4. IDENTIFIED AREAS

We have identified several areas to conduct future research in the field. As the traditional machine learning approaches have handcrafted features extraction method, these methods were not robust for diabetic retinopathy task. The limitations in traditional hand-crafted features were overcome by the introduction of Convolutional Neural Networks (CNNs), the problem of hand-crafted feature extraction was mitigated with more robust automated feature extraction method. Further, there is a scope of the future of research to make deep learning algorithms more accurate and efficient.

As the prevalence of diabetes mellitus is drastically increasing [26], there is a need for early detection and treatment of Diabetic Retinopathy which has a capability to remarkably diminish the odds of permanent vision loss. The integration of deep learning algorithms with circulating biomarkers [27] may help in the early detection of diabetic retinopathy.

The current Diabetic Retinopathy screening programs typically utilize retinal fundus photography [28], which depends on the manual human readers for the evaluation of reports. This screening method is arduous and prone to inconsistency. Moreover, the current screening systems do not aid the peripheral retina and require pharmacological pupil dilation. So, there can be a scope of another retinal screening method in the future work to mitigate the problems of fundus photography. The possible substitute can be Thermal Imaging modality, which showed its competency for breast cancer detection, diabetic foot, and other eye diseases detection.

Furthermore, the blood vessels extraction is a challenging task considering the different Optic Nerve Hypoplasia structure. Therefore, there is not a single technique that can cover all these problems. There is still a need to propose more efficient algorithms for the identification of Diabetic Retinopathy detection related structure and retinal changes. Modification in health care delivery standards will also be required, along with new diagnostic methods to match the increasing prevalence of diabetic retinopathy.

## 5. CONCLUSION

As the prevalence of Diabetic Retinopathy is predicted to increase, there is a need to find an efficient means for Diabetic Retinopathy detection. The early detection of Diabetic Retinopathy can significantly help to avoid irreversible vision loss. The capability of computer-aided detection systems to accurately and efficiently detect diabetic retinopathy can be the most suitable solution to the Diabetic Retinopathy problem. Despite significant advances in the computer-aided Diabetic Retinopathy detection over the past ten years, certain factors need to be addressed in future research like advancement in deep learning algorithms, integration of deep learning with circulating biomarkers, thermal imaging modality to overcome limitations of fundus photography, study on the blood vessels extraction, and better health care delivery standards. The major focus of this study is to direct the researchers about the limitations of current methods and identify the specific areas in the field to boost future research.

## BIOGRAPHIES

| | |
|---|---|
| 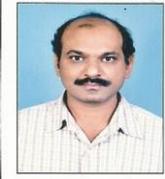 | Dr. P. S. Prasad is working as Professor and Head Computer Science Department at Priyadarshini Institute of Engineering and Technology, Nagpur, Maharashtra, India. He holds PhD degree from 2014. He has published more than 50 papers in National and International Journals and Conferences. He is a Reviewer for few IEEE Transactions and International Journals. He is Member of IEEE, IACSIT and various professional bodies which are of National and International repute. He is having more than 21 Years of experience. |
| 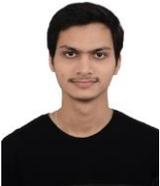 | Mr. Saket Chaturvedi is an undergraduate student (2016-2020) at the Computer Science Department of Priyadarshini Institute of Engineering and Technology, India. He has worked on the computer-aided diagnosis of skin cancer employing Deep Learning techniques. Since 2016, he has explored several healthcare projects in the Kaggle competitions. |
| 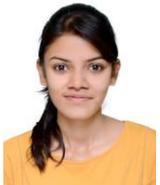 | Ms. Kajol Gupta is an undergraduate student (2016-2020) at the Computer Science Department of Priyadarshini Institute of Engineering and Technology, India. Her interest lies in the applications of Machine Learning and Deep Learning. |